\documentclass{article}
\pdfoutput=1

\usepackage{arxiv}
\usepackage{graphicx}
\usepackage{subfigure}
\usepackage[utf8]{inputenc} 
\usepackage[T1]{fontenc}    
\usepackage{hyperref}       
\usepackage{url}            
\usepackage{booktabs}       
\usepackage{amsfonts}       
\usepackage{nicefrac}       
\usepackage{microtype}      
\usepackage{lipsum}
\usepackage[numbers]{natbib}
\usepackage{amsmath}
\usepackage{authblk}

\title{Neuroevolution of a Recurrent Neural Network for Spatial and Working Memory in a Simulated Robotic Environment}

\author[1,*]{\textbf{Xinyun Zou}}
\author[2]{\textbf{Eric O.~Scott}}
\author[3]{\textbf{Alexander B.~Johnson}}
\author[4]{\textbf{Kexin Chen}}
\author[3]{\textbf{Douglas A.~Nitz}}
\author[2]{\textbf{Kenneth A.~De Jong}}
\author[4,1]{\textbf{Jeffrey L.~Krichmar}}

\affil[1]{Department of Computer Science, University of California, Irvine, Irvine, CA 92697, USA}
\affil[2]{Department of Computer Science, George Mason University, Fairfax, VA 22030, USA}
\affil[3]{Department of Cognitive Science, University of California, San Diego, La Jolla, CA 92093, USA}
\affil[4]{Department of Cognitive Sciences, University of California, Irvine, Irvine, CA 92697, USA}
\affil[*]{Correspondence Email: xinyunz5@uci.edu}

\begin{document}
\maketitle

\begin{abstract}
Animals ranging from rats to humans can demonstrate cognitive map capabilities. We evolved weights in a biologically plausible recurrent neural network (RNN) using an evolutionary algorithm to replicate the behavior and neural activity observed in rats during a spatial and working memory task in a triple T-maze. The rat was simulated in the Webots robot simulator and used vision, distance and accelerometer sensors to navigate a virtual maze. After evolving weights from sensory inputs to the RNN, within the RNN, and from the RNN to the robot's motors, the Webots agent successfully navigated the space to reach all four reward arms with minimal repeats before time-out. Our current findings suggest that it is the RNN dynamics that are key to performance, and that performance is not dependent on any one sensory type, which suggests that neurons in the RNN are performing mixed selectivity and conjunctive coding. Moreover, the RNN activity resembles spatial information and trajectory-dependent coding observed in the hippocampus. Collectively, the evolved RNN exhibits navigation skills, spatial memory, and working memory. Our method demonstrates how the dynamic activity in evolved RNNs can capture interesting and complex cognitive behavior and may be used to create RNN controllers for robotic applications.
\end{abstract}


\section{Introduction}

The cognitive map, a concept raised by Edward C. Tolman in 1930s \cite{tolman1948cognitive}, describes that the mental representation of a physical space could be built by integrating knowledge gained from the environmental features (e.g., goals, landmarks, and intentions). We used the Webots robot simulation environment \cite{michel2004cyberbotics} to investigate cognitive map behavior observed in rats during a spatial and working memory task, known as the triple T-maze \cite{olson2017subiculum,olson2020secondary}. We suggest that similar behavior could be observed in a robot that had a biologically plausible neural network evolved to solve such a task. In this task, the rat or the robot must take one of four paths to receive a reward. If it repeated a path, there would be no additional reward. It would eventually learn to quickly reach each of the four rewards with minimal repeats. This requires knowledge of where it is now, where it has been, and where it should go next. 

In our Webots setting, we designed a 3-D environment that resembled the rat experiment. The proximity sensors, the linear accelerometer and the grayscale camera pixels of a simulated e-puck robot \cite{mondada09theepuck} provided sensory input for a recurrent neural network (RNN). The RNN output directly manipulated the motor speed of the e-puck. Using neuroevolution, the input weights into the RNN, the recurrent weights within the RNN, and the output weights from the RNN were evolved based on an objective designed to replicate the rat behavior. 

Our results show that the evolved RNN was capable of guiding the robot through the triple-T maze with similar behavior to that observed in the rat. Our analysis of the RNN activity indicated that the behavior was not dependent on any one sensory projection type but rather relied on the evolved RNN dynamics. Furthermore, the population of neurons in the RNN were not only sufficient to predict the robot's current location but also carried a predictive code of future intended reward paths. Furthermore, the present method for evolving neural networks for robot controllers may be applicable to other memory tasks.

\section{Methods}
\label{method}

\begin{figure}
\centering
\includegraphics[width=0.6\textwidth]{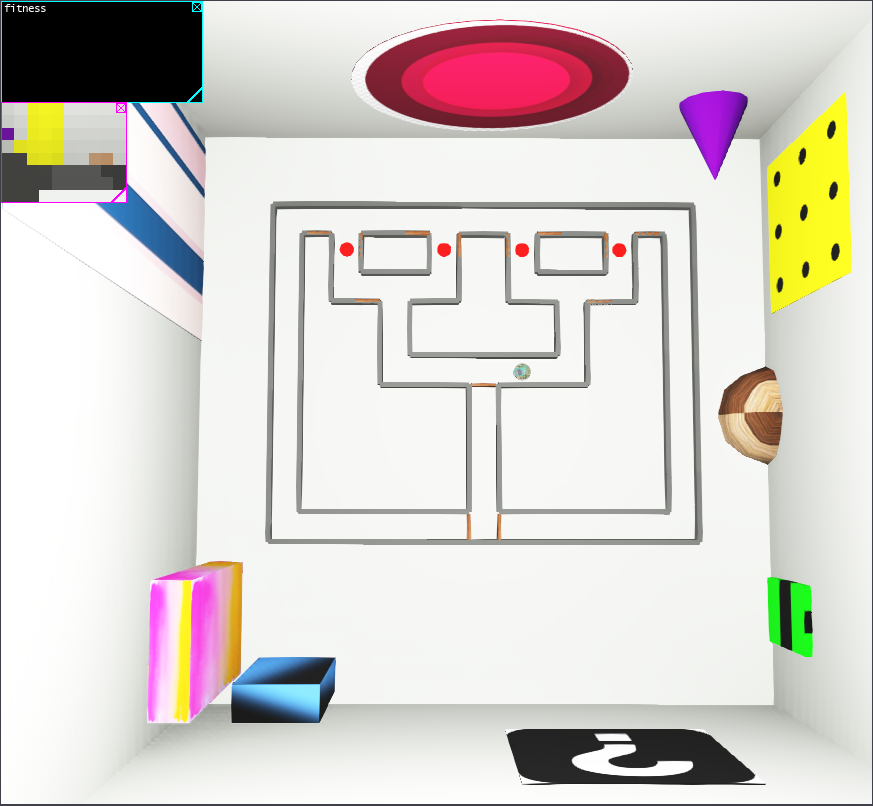}
\caption{The maze visualization in Webots. As shown in the figure, there were corridors that the robot could traverse and landmarks on the wall. The e-puck robot is denoted by the small green circle. The red circles denote the reward locations. Note that these rewards were not visible to the robot's sensors.}
\label{fig:landmarks}
\end{figure}

\begin{figure}
\centering
\includegraphics[width=0.25\textwidth]{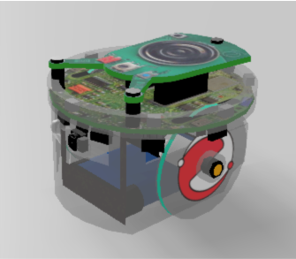}
\caption{The 3D simulated e-puck robot. The picture is adapted from \citet{simulatedepuck}.}
\label{fig:e-puck}
\end{figure}

\begin{figure}
\centering
\includegraphics[width=0.45\textwidth]{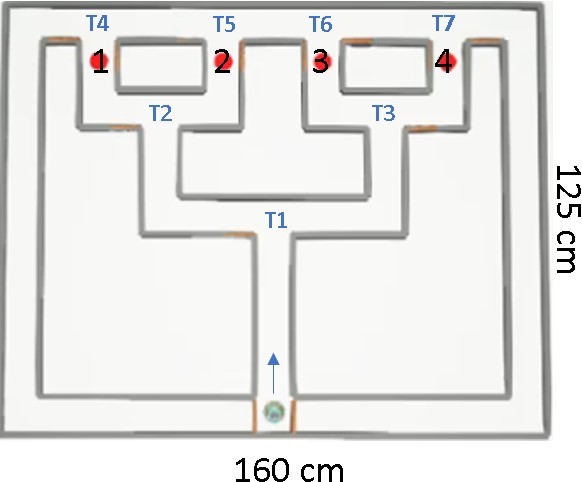}
\caption{A closer look of the maze, with 4 rewards labeled in red circles and 7 T-intersections labeled in blue font. Home was located at the e-puck's current position (bottom-middle) in this figure. }
\label{fig:maze}
\end{figure}

We picked Webots \cite{michel2004cyberbotics} as our virtual robotic environment. Inside this 3D simulator, a triple T-maze was constructed that closely followed the dimensions and landmarks used in the rat experiment \cite{olson2017subiculum}. Figure \ref{fig:landmarks} shows the maze simulation environment. The red circles, which denote the location of the rewards, were not observable by the robot and are only included in the figure for illustrative purposes. The agent was an e-puck robot \cite{mondada09theepuck} which has an accelerometer, a front camera, 8-direction proximity sensors, several LEDs, and 2 wheel motors (see Figure \ref{fig:e-puck}). The e-puck needed to learn by neuroevolution to find four rewards (and return home after each reward visit) with minimal repeats before timeout. Its actuation was updated every 64 milliseconds. The timeout threshold was tuned to 5000 steps (i.e., 320 seconds in real-time) per trial to guarantee enough time to visit all four rewards with minimal repeats and some tolerance for slight movement variations. For each trial, the robot would always start from the home position in the bottom middle part of the maze and move upward (see the e-puck's location in Figure \ref{fig:maze}). After reaching a T-intersection, a door behind the robot would close to prevent backtracking. Since the robot could only move forward in a reward path, it could neither revisit places on the same path nor switch to a different one before completion of the previous path. After the robot moved within 6 cm from the center position of a novel reward in a trial, the reward was added to the objective function given by Equation \ref{eq:fitness1}. After the robot passed through the third T-intersection right above any reward position, an additional door on the right/left would close to enforce its usage of the closer return path to home before exploring the next reward path. It should be noted that although the doors prevented backtracking, the robot still needed to evolve its ability to move smoothly and efficiently through the corridors to receive all four rewards with minimal repeats prior to the timeout.

The rotation speed of an e-puck was ranged between -3.14 rad/s and 6.28 rad/s. The evolved output weights in the RNN adjusted the speed and the turning rate of the robot to navigate the task with optimal and stable performance. To prevent from being stuck at corners or T-intersections, the robot had a default obstacle avoidance algorithm that used the 8 proximity sensors, which only influenced movement when the robot was very close to an obstacle, to move away from the closest point of contact \cite{fajen2003behavioral}. The obstacle avoidance motor signal was added to the rotational speed of the motors dictated by the RNN output.

We conducted 5 evolutionary runs and selected the best performing agent from each run for further analysis. An evolutionary run was composed of 200 generations to achieve optimal performance. During each generation, there were 50 genotypes generated according to the evolutionary algorithm described in Section \ref{sec:ea}. For each genotype during the evolutionary process, the fitness value was recorded as an average over 5 trials to improve robustness in the selection. In each test scenario afterwards, each of the 5 best performing agents from these runs was utilized to run 20 demo trials with the same task setting and timeout threshold. For each demo trial, the activities of 50 recurrent neurons and the robot positions were recorded at each time step for further analysis.

\subsection{Network Architecture}
\label{sec:architecture}

\begin{figure*}
\centering
\includegraphics[width=0.9\textwidth]{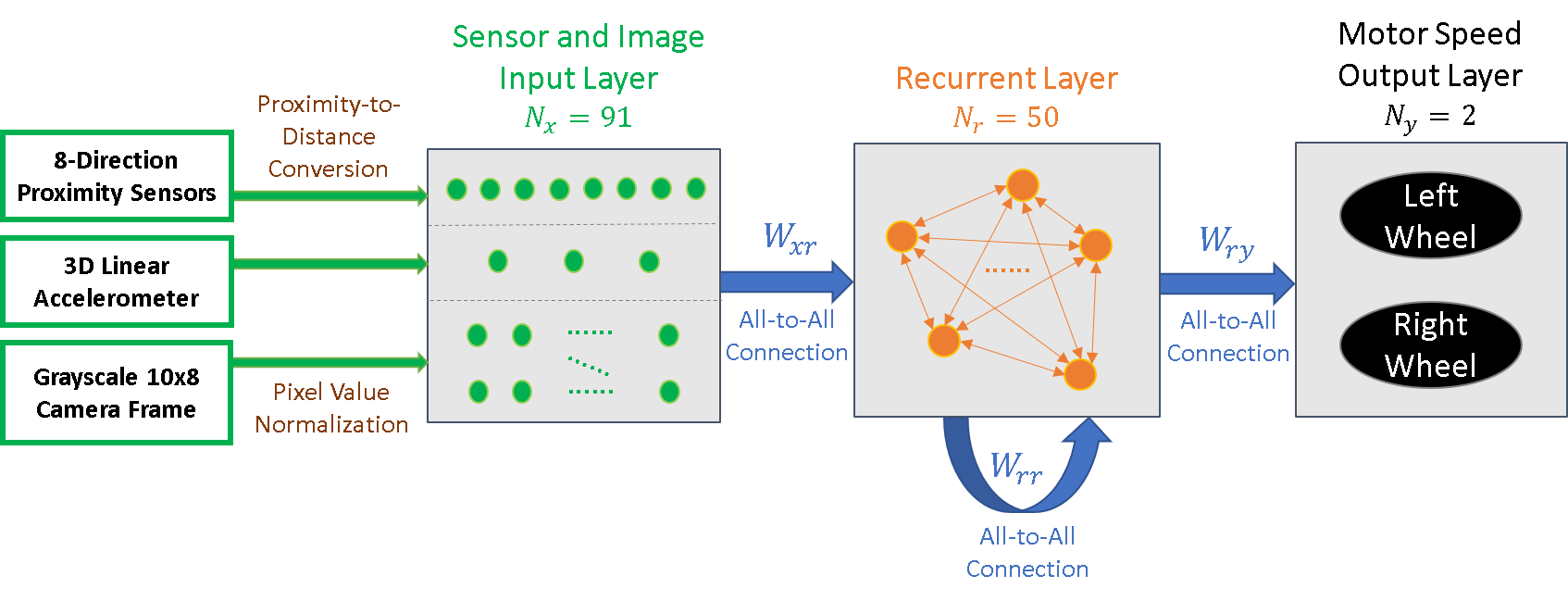}
\caption{The neural network architecture for controlling the e-puck robot in Webots. Sensors were converted into input neural activities. The input weights ($W_{xr}$), recurrent weights ($W_{rr}$), and output weights ($W_{ry}$) were evolved concurrently. The output weights dictated the left and right rotational wheel speed of the e-puck.}
\label{fig:architecture}
\end{figure*}

The neural network architecture received inputs from the e-puck's 8-direction proximity sensors, 3D linear accelerometer values, and normalized pixel values from its $10\times 8$ grayscale camera frame (Figure \ref{fig:architecture}). These 91 input neurons were fully connected to 50 recurrent neurons, which were fully connected with one another. This recurrent layer was then fully connected with the two neurons in the output layer that controlled the rotational speed of the two wheel motors separately. 

\subsubsection{Recurrent neural network}
For each recurrent neuron $i$, its recurrent activity $R_i$ was updated at every time step $t$ according to Equation \ref{eq:recur}:
\begin{equation}
 \left.\begin{aligned}
        & R_i^{(t=0)} = 0.0,\\
        & \text{synIn}_i^{t} = \sum_{k \in in} W_{ki}\cdot x_k^{t} + \sum_{j \in rec}^{j\neq i} W_{ji}\cdot R_j^{(t-1)},\\
        & R_i^{t} = (1-p)\cdot \tanh{\left(\text{synIn}_i^{t}  \right)} + p \cdot R_i^{(t-1)}
       \end{aligned}
 \right\}
 \label{eq:recur}
\end{equation}
Here $x$ denoted the input sensor value and $W_{ki}$ was the weight from Neuron $k$ to Neuron $i$. In other words, the synaptic input for each recurrent neuron ($\text{synIn}_i$) contained (1) the summation of the product of each sensor value and the corresponding input weight plus (2) the summation of the product of every other recurrent neuron's previous activity and the corresponding recurrent weight. The $\tanh$ wrap ensured the recurrent activity between -1.0 and 1.0. A small $p$ value of 0.01 helped to avoid any abnormal performance of the computed recurrent activity. The recurrent activity was then used to compute the rotational speed of each wheel motor by multiplying with the output weight.

\subsubsection{Evolutionary algorithm}
\label{sec:ea}

An evolutionary algorithm was used to evolve the input, recurrent, and output weights ($W_{xr}$, $W_{rr}$, and $W_{ry}$ in Figure \ref{fig:architecture}). Because of all-to-all connections, there were a total of 7150 genes for each genotype. The fitness value of each genotype was the average over 5 trials. The evolutionary algorithm used a population of 50 genotypes selected by linear ranking. Two-point crossover and mutation (with a decaying mutation standard deviation) were applied to reproduce the non-elite 90\% of the population. The mutation rate was 0.06 and the mutation standard deviation decreased throughout a run via the function: 
\begin{equation}
\text{mutation\_std} = 0.3\times \frac{50.0}{50.0+m}
\label{eq:mutstd}
\end{equation}
Here $m$ denotes the generation index. 

The fitness function is shown in Equation \ref{eq:fitness1}:
\begin{equation}
\text{fitness} = \text{ num\_obtainedRwds} + \text{portion\_routes\_completed} {-} 0.2 \times \text{num\_repeats}
\label{eq:fitness1}
\end{equation}
During each trial, we would reward (1) each non-repetitive visit of any reward arm (i.e., $0\sim 4$) and (2) the portion of reward path visits for which home was returned afterwards (i.e., $0\sim 1$); meanwhile, we would penalize every repeated visit.

\subsection{Bin-based Recurrent Activities}
\label{sec:method-bin-analysis}

To analyze the performance of the robot after evolution, we divided the 1.6m-by-1.25m maze into 0.08m-by-0.10m sized bins, which was close to the e-puck’s diameter (0.074m) (Figure \ref{fig:bins}). We computed the average activity of each recurrent neuron for each bin in the maze. For each of the 5 best performing agents (from 5 evolutionary runs), we used 15 demo trials to generate the expected bin-based activity matrix and 5 demo trials for bin occupancy prediction. The results are shown in Sections \ref{sec:results_spatial} and \ref{sec:results_prospective}. 

\begin{figure}
\centering
\includegraphics[width=0.5\textwidth]{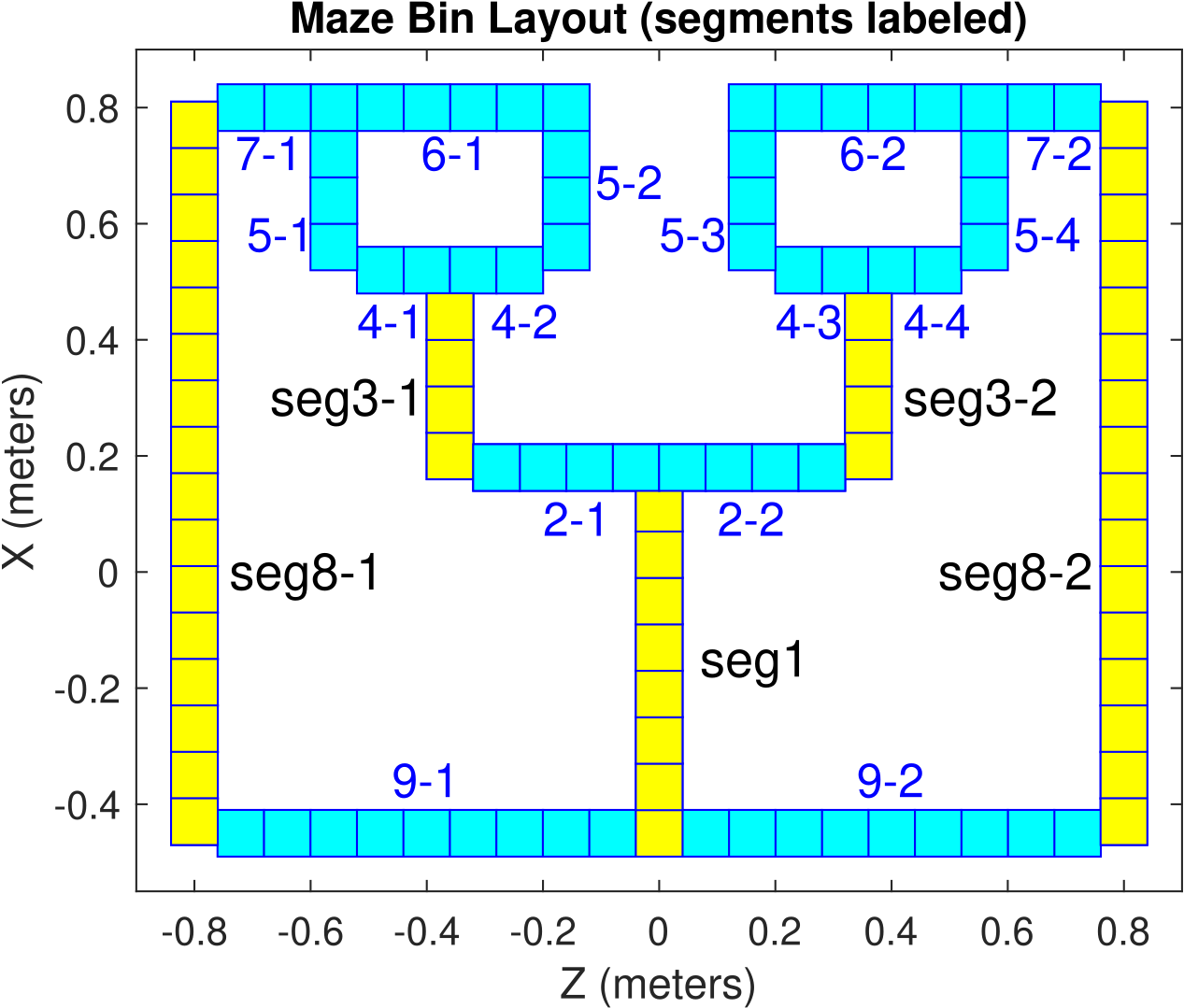}
\caption{The maze layout with 110 bins of size 0.08m-by-0.10m. Segments used for the trajectory-dependent coding analysis are labeled as seg1, seg3-1, seg3-2, seg8-1 and seg8-2 in yellow. }
\label{fig:bin_segments}
\end{figure}

\subsubsection{Spatial memory analysis}
We used the RNN activity to predict the robot's location in the maze. Since there were 50 recurrent neurons, for each demo trial, we generated a bin-based recurrent activity matrix of size $110\times50$. The activity at each bin for each neuron was averaged over all steps spent on that bin in a trial. Then we obtained an expected bin-based activity matrix of size $110\times50$ by taking the average over all the matrices for the first 15 demo trials. The remaining 5 trials were used to analyze the RNN's ability to encode location. For each of these 5 test trials, we compared each bin's RNN activity vector, which had a length of 50, with all 110 expected bin-based activity vectors using a Euclidean distance metric. The predicted bin was the smallest Euclidean distance to that actual bin. Thus, the Euclidean distance denotes the prediction error in bins. For example, if the Euclidean distance was 6, there was a perfect prediction error of 6 bins or approximately 0.5m (see Figure \ref{fig:bin_occupancy}). 

\subsubsection{Trajectory-dependent analysis}
After traversing some of the maze's vertical (South-to-North) segments, the robot would decide to turn left or right at a T-intersection. We analyzed if the RNN activity during traversal of a vertical segment could predict the robot's future path or the robot's prior path. As shown in Figure \ref{fig:bin_segments}, Segment 1 is the vertical segment right before the first T-intersection on the path for any of the four rewards. Segment 3-1 (or Segment 3-2) denotes the vertical segment right before the second T-intersection on the path for Reward 1 or 2 (or for Reward 3 or 4). Segment 8-1 (or Segment 8-2) represents the vertical segment for returning from Reward 1 or 2 (or from Reward 3 or 4). Similar to the method described above, we computed an expected bin-based activity matrix for each of these segments from the first 15 demo trials for each reward path. We then used the remaining 5 demo trials to test whether the RNN activity could predict which path the robot was taking (i.e., \textit{Prospective}; seg1, seg3-1, seg3-2) or which path the robot was returning from (i.e., \textit{Retrospective}; seg8-1 and seg8-2).

\section{Results}

\subsection{Evolutionary Runs}
\label{sec:results_evolve}

\begin{figure}
\centering
\includegraphics[width=0.95\textwidth]{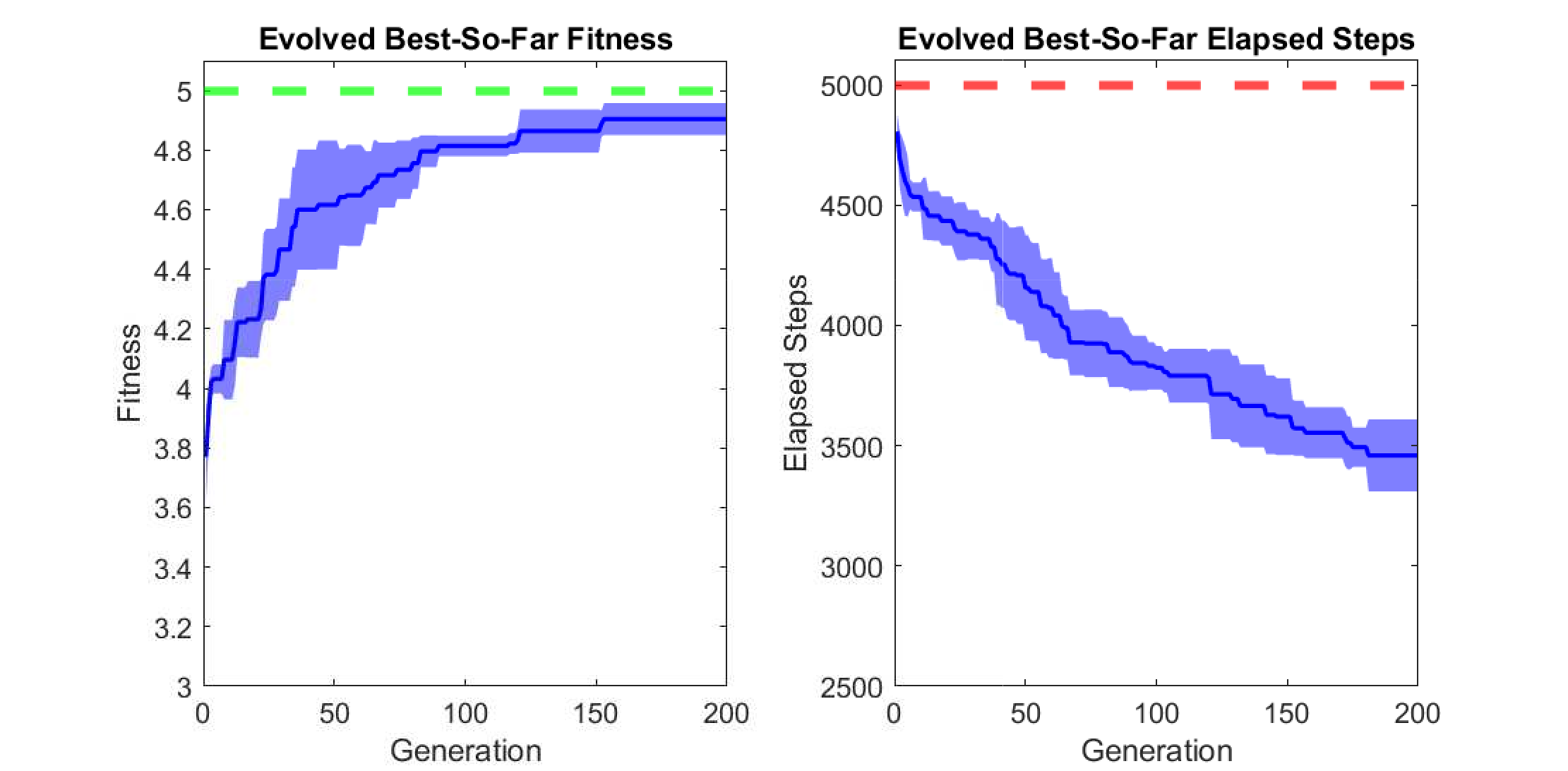}
\caption{The evolutionary performance (left: fitness, right: number of elapsed time steps) for the best-so-far agent evolved in each generation. Each subplot was averaged over 5 runs with 200 generations per run. The shaded area denotes the 70\% confidence level. }
\label{fig:evolved_trend}
\end{figure}

\begin{figure}
\centering
\includegraphics[width=0.6\textwidth]{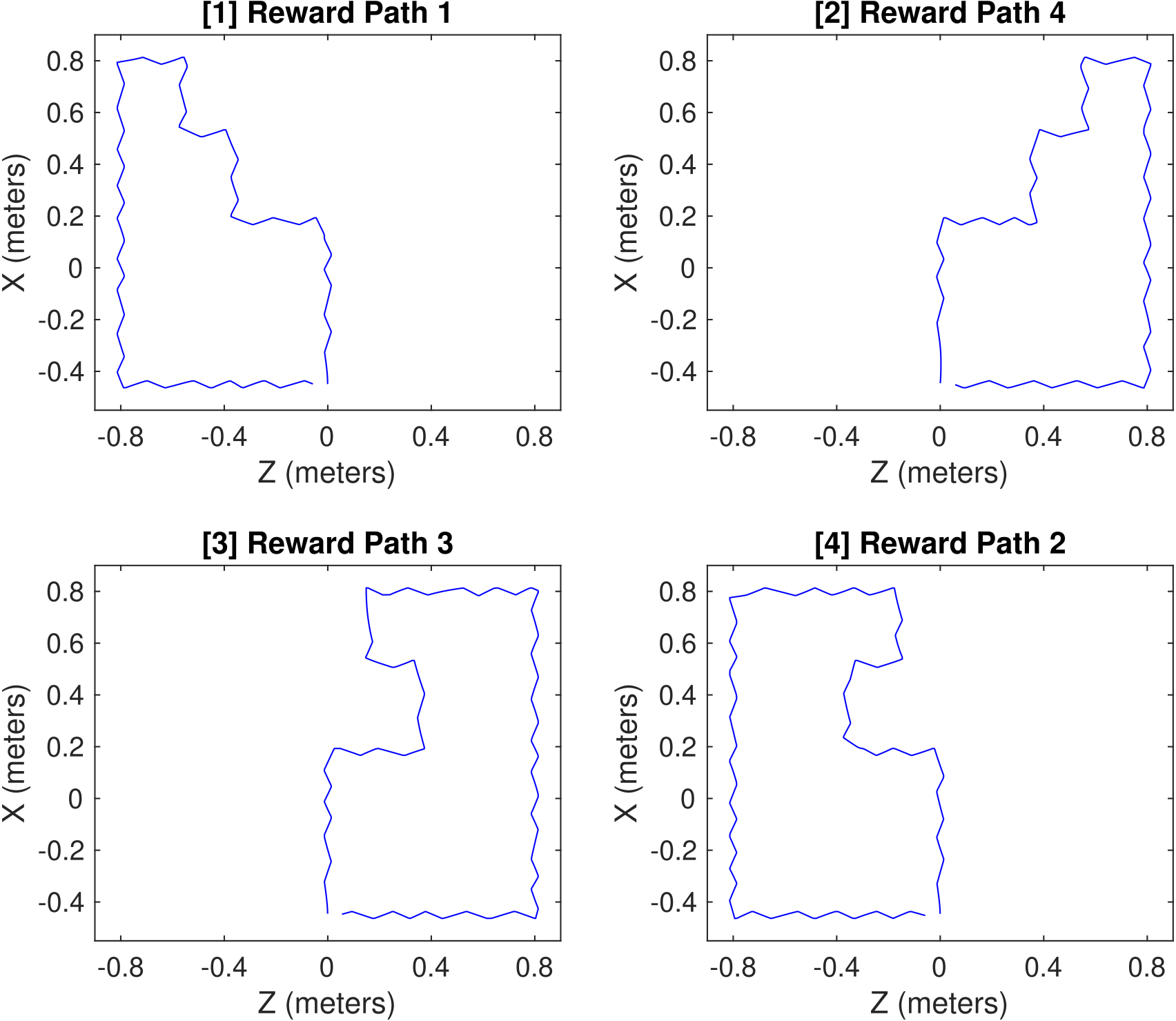}
\caption{The trajectory of a perfect trial, which covered reward paths 1, 4, 3, 2 in order with no repeated visits of any path. }
\label{fig:demo_trajectory}
\end{figure}

Successful behavior, similar to that observed in rats, emerged from the evolutionary process. We ran all simulations using Webots (version R2020a) on a desktop with one GPU (Nvidia GeForce GTX 1080 Ti). Figure \ref{fig:evolved_trend} shows the best-so-far evolutionary performance and the number of elapsed steps for five runs. Each run lasted 200 generations, with 50 genotypes per generation. In each generation, the fitness value of a genotype was averaged over 5 trials. By the end of each run, the best-so-far fitness curve reached a plateau close to the maximal fitness value of 5, whereas the number of steps taken to complete the task dropped below 3500 steps per trial on average. An example perfect trial trajectory (with no repeated visits of any reward path and a fitness of 5) can be observed in Figure \ref{fig:demo_trajectory}.

\subsection{Ablation Performance}
\label{sec:results_ablation}

\begin{table}
  \caption{Ablation performance (mean $\pm$ the 95\% confidence level) over 20 trials per ablation test for the best performing agent in each of the 5 evolutionary runs. The values highlighted with bold fonts and asterisks denote ablations that had a significant impact on the performance. Significance threshold was a p-value < 0.01/6 = 0.0017 using the Wilcoxon Rank Sum test.}
  \label{tab:ablation}
  \centering
  \begin{tabular}{l|l|l}
    \toprule
    & Fitness & Elapsed Steps\\
    \midrule
    No Ablation & 3.65$\pm$0.14 & 4644$\pm$169 \\
    Proximity Sensors & 3.29$\pm$0.16 & 4774$\pm$128 \\
    Linear accelerometer & 3.49$\pm$0.16 & 4640$\pm$131 \\
    Grayscale Vision & 3.21$\pm$0.17 & 4785$\pm$67 \\
    \textbf{Input Weights} & 2.93$\pm$0.085 & \textbf{4981$\pm$16$\ast$} \\
    \textbf{Recurrent Weights} & \textbf{2.95$\pm$0.11$\ast$} & \textbf{4946$\pm$43$\ast$} \\
    \textbf{Output Weights} & \textbf{2.84$\pm$0.10$\ast$} & \textbf{4978$\pm$22$\ast$} \\
  \bottomrule
\end{tabular}
\end{table}

We carried out a set of ablation simulations to test whether performance was dependent on any sensory projection type or just evolved weights in the neural network (Figure \ref{fig:architecture}). To test this, we either shuffled different sensor input values or shuffled the RNN input ($W_{xr}$,) recurrent ($W_{rr}$), or output ($W_{ry}$) weights. For each of the 6 ablations, we ran the best performing agent from each of 5 evolutionary runs in demo trials. The results were averaged over 20 demo trials for each shuffle test. Random shuffle sequences occurred at each time step for each demo trial.

The ablation studies show that the dynamics of the RNN was critical for performance (Table \ref{tab:ablation}). We compared the control (no ablation) with the 6 ablation groups. Since there were 6 comparisons, the significance threshold for the p-value is 0.01/6 = 0.0017 based on a Bonferroni correction. Interestingly, none of the sensory projection ablations had a significant impact on performance. However, ablating the evolved weights (input, recurrent, and output) all had a significant impact. That suggests that it was the recurrent neural network dynamics that were key to performance. Moreover, that performance was not dependent on any one sensory projection type.

\subsection{Order Effects}
\label{sec:results_order}

\begin{figure}
\centering
\includegraphics[width=0.68\textwidth]{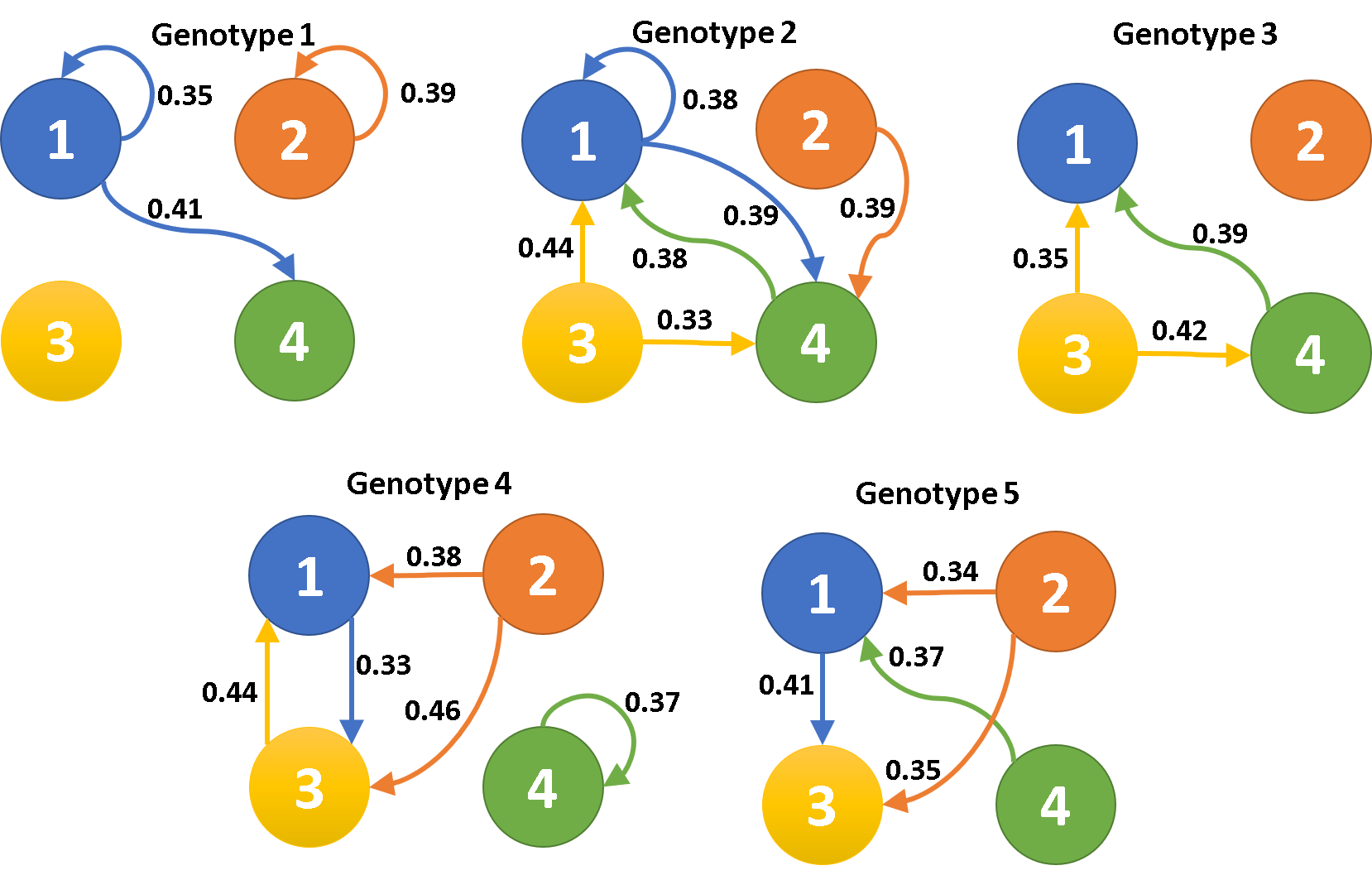}
\caption{The trend of transitioning from one reward path to the next was different for each best performing agent (with a different genotype evolved) after 200 generations for each of the five evolutionary runs. The numbered circles denote the reward path, and the labeled arrows denote the probability of transitioning from one reward path to another.}
\label{fig:transition}
\end{figure}

We wanted to test if the robot evolved strategies to solve the triple-T maze task. Rats tend to show idiosyncratic behavior in the same maze setting \cite{olson2020secondary,olson2021complementary}. For instance, a rat alternated by going to the left side of the maze towards Rewards 1 and 2, and then the right side of the maze towards Rewards 3 and 4. Idiosyncratic behavior did emerge in our evolved robots. Although we did not observe the robots alternating between sides of the maze, each best performing agent for an evolutionary run exhibited a unique strategy for traversing the maze. Figure \ref{fig:transition} shows the probability of transitioning from one reward path to the next. Only transitions probabilities that are greater than 0.33 are shown. We did find some generalities between genotypes. For example, the agents evolved in Genotypes 2, 3, and 5 tended to transition from the path for Reward 4 to the path for Reward 1. The agents evolved in Genotypes 2, 3, and 4 tended to transition from the Reward 3 path to the Reward 1 path. In both cases, the robot was navigating the right side of the maze before transitioning to the left side of the maze. These strategies that emerged in our robot and in the rat may simplify the task by breaking down the problem into chunks (e.g., first go to the right, and then go to the left).

\subsection{Spatial Coding in the RNN}
\label{sec:results_spatial}

We investigated if the RNN activity was sufficient to predict the robot's position. If the activities of the 50 recurrent neurons could accurately encode the position, then the robot might be using this piece of information to solve the maze task. Borrowing techniques from neuroscience \cite{olson2017subiculum, Wilson1993, ferbinteanu2003prospective}, we tested whether the RNN contained spatial information with a population code.

\begin{figure}
\centering
\includegraphics[width=0.6\textwidth]{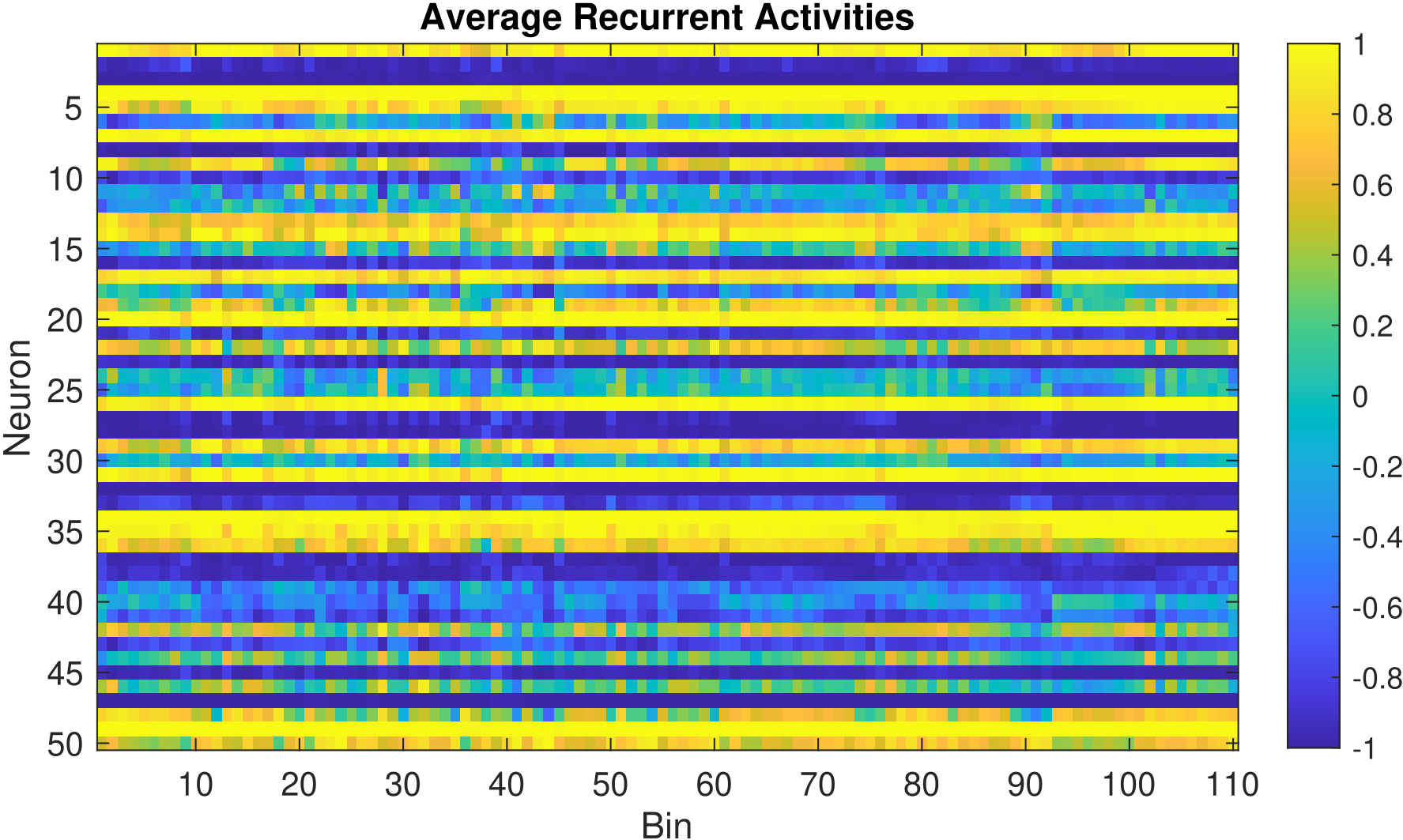}
\caption{The average bin-based activities for all 50 recurrent neurons on 110 bins across the entire maze for a top performing agent.}
\label{fig:bins}
\end{figure}

Individual neurons in the recurrent layer did not seem to have place information. For example, Figure \ref{fig:bins} shows the average bin-based recurrent activities across the entire maze from 15 demo trials of a top performing agent. Each bin's activity per trial divided by the number of time steps spent on that bin. The activity of each neuron is noisy with some neurons being highly active, quiescent, or oscillating. 

\begin{figure}
\centering
\includegraphics[width=0.68\textwidth]{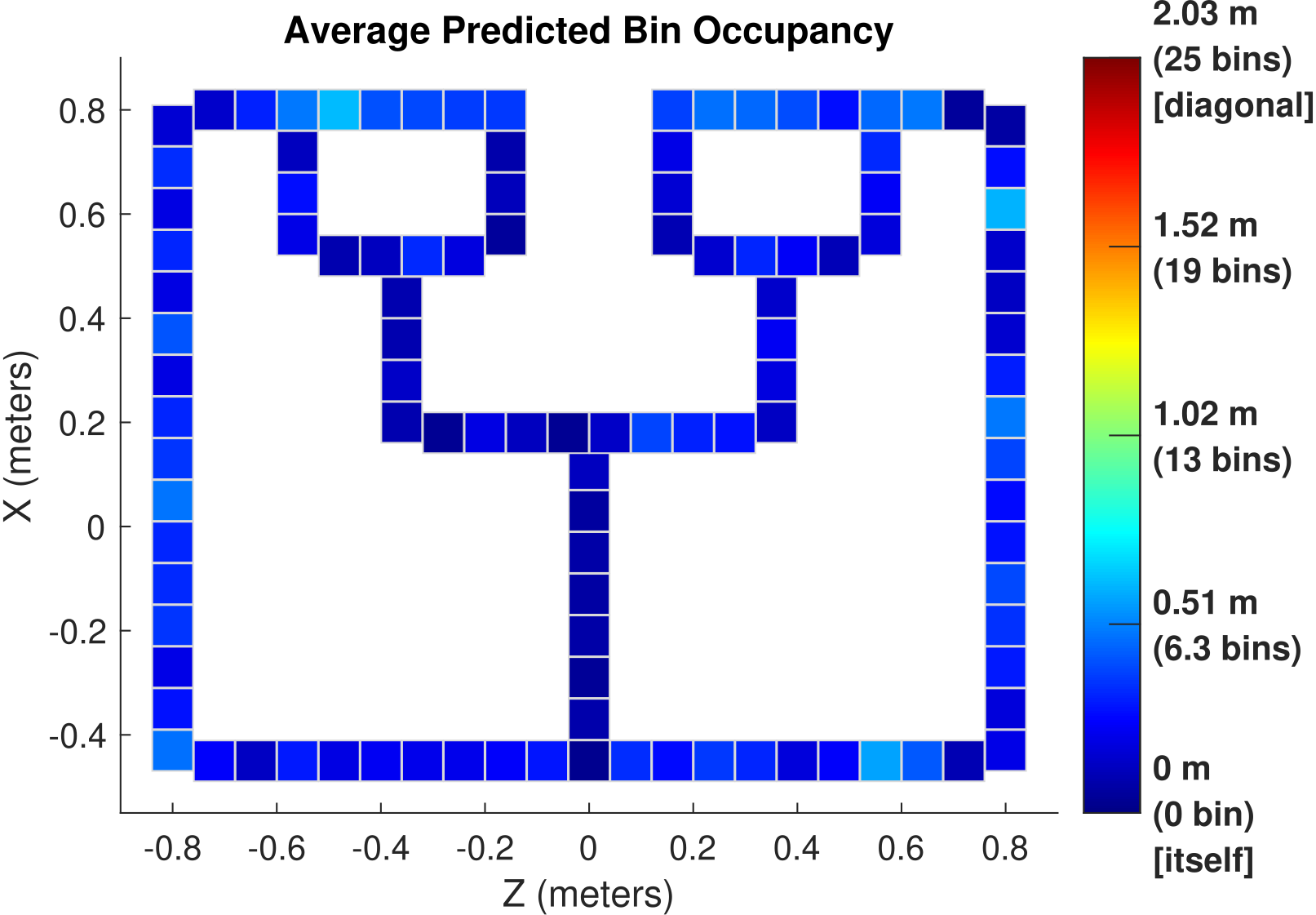}
\caption{The average predicted bin occupancy over all 25 test trials of the best performing agent from each of the five evolutionary runs. A dark blue bin (if existing) would have perfect prediction right at itself (distance = 0), whereas a dark red bin (if existing) would have a farthest prediction (across the diagonal of the entire maze). }
\label{fig:bin_occupancy}
\end{figure}

However, the population of 50 RNN neurons was able to predict the robot's location throughout the maze. With the method described in Section \ref{sec:method-bin-analysis}, Figure \ref{fig:bin_occupancy} shows the location prediction for each bin in all 25 test trials (with 5 best performing agents from 5 evolutionary runs and 5 trials per agent). It is apparent from the figure that the RNN activity was sufficient to predict the robot's position in the maze. The robot's position was predicted with perfect accuracy on 58\% of the bins, and the predicted error had an average distance of 3.1 bins (i.e., 0.25 meters).

\subsection{Trajectory-Dependent Coding in the RNN}
\label{sec:results_prospective}

Solving the triple-T maze task requires the agent, whether it is a robot or an animal, to remember which path it has already taken, as well as to decide which path to take next. We hypothesized that the dynamic activity of the RNN carried such information, which is known as retrospective coding (i.e., where it has been) and prospective coding (i.e., where it intends to go).

To test whether there was retrospective coding and/or prospective coding in the RNNs, we analyzed the RNN's ability to encode trajectory-dependent information at a population level. Table \ref{tab:prospective} shows how well the RNN could predict the robot's future path based on the activity of Segments 1, 3-1 and 3-2, and how well the RNN could predict the robot's past path based on the activity of Segments 8-1 and 8-2. The probability of correct path prediction on Segment 1, where the robot could take one of 4 paths, was well above chance level (t-test; p < 0.0001). The correctness on Segment 3-1 or 3-2, where the robot could take one of two paths, was also well above chance level (t-test; p < 0.0001 for Segment 3-1 and p < 0.005 for Segment 3-2). This suggests that the RNN carried a prospective code of where the robot intent to go next. The probabilities of correct path prediction on Segments 8-1 and 8-2 were not above chance (t-test > 0.05), which indicates they did not predict if the robot came from one of the 2 prior paths. 

Taken together, these results suggest that the evolved RNN had prospective information in that it could be discerned which direction the robot would take before turning left and right. It is interesting that we were not able to observe retrospective information in the RNN population since some knowledge of where the robot had already visited was necessary for the observed performance.

\begin{table}
  \caption{Average prospective path prediction for different segments. }
  \label{tab:prospective}
  \centering
  \begin{tabular}{l|ccccc}
    \toprule
    &Seg1 & Seg3-1 & Seg3-2 & Seg8-1 & Seg8-2\\
    \midrule
    correctness & 41\% & 77\% & 69\% & 47\% & 55\% \\
    bins off & 0.9 & 0.2 & 0.2 & 2 & 2 \\
  \bottomrule
\end{tabular}
\end{table}

\section{Discussion}

The present work demonstrated that a robot controlled by an evolved RNN could solve a spatial and working memory task where the robot needed to navigate a maze and remember not to repeat paths it had taken previously. The RNN population activity carried spatial information sufficient to localize robot, and the RNN population activity carried predictive information of which path robot intended on taking. 
Behavior was dependent on RNN dynamics and not any particular sensory channel. The present method shows that complex robot behavior, using a detailed robot simulation, could be realized by evolving all weights of a RNN.

\subsection{Evolved RNN with Spatial and Working Memory}

We evolved a RNN to control a robot in a spatial and working memory task that replicated behavior and neural activities observed in rats \cite{olson2017subiculum}. The robot was able to navigate the triple T-maze efficiently by reaching all four rewards with minimal repeats. Successful performance required the robot to have spatial knowledge and working memory of which rewards it had already visited. Prospective information, in which RNN activity predicted the robot's intention, emerged in the simulations. Although not observed in the analysis, the neural network must have had retrospective information to minimize repeating previously traversed paths. 

At the population level, the evolved RNN activity has similar characteristics as the hippocampus. It has been observed that the population activity of the CA1 region in the hippocampus can accurately predict the rat's location in a maze \cite{Wilson1993,olson2021complementary}. Furthermore, journey-dependent CA1 neurons have been observed in the rat that can predict the upcoming navigational decision \cite{ferbinteanu2003prospective}. Similar to CA1, the RNN received speed, direction, and visual information as input, and combined these types of sensory information to construct a journey-dependent place code \cite{taube1990head,sargolini2006conjunctive,kropff2015speed,o1976place,hafting2005microstructure,sun2019ca1,potvin2007contributions,frost2020anterior}.

In \citet{olson2021complementary} and in other rodent studies, it has been observed that rats acquire individual strategies to solve navigational tasks. For example, in the triple-T task, many rats alternated between the left and right arms of the maze. Similarly, our robot demonstrated this idiosyncratic behavior. Each best performing agent for an evolutionary run had an order-dependent pattern for taking different paths in the maze. It suggests that the RNN evolved a strategy breaking down the complex maze task into simpler pieces, which may also be how animals solve tough problems.

\subsection{Behavioral Dependence on RNN Dynamics}
To investigate the dependence of the robot behavior on different components of the RNN neural architecture, we conducted an ablation study (see Figure \ref{fig:architecture}). Results in Section \ref{sec:results_ablation} show that ablating a given sensory channel had no significant impact on performance. However, ablating the evolved weights (input, recurrent, and output) all had a significant impact. This suggests that it was the recurrent neural network dynamics that were key to performance, and that performance was not dependent on any one sensory projection type. The results justify evolving all weights in the RNN structure, rather than evolving only the readout weights as is often done in Liquid State Machines (LSM) or Echo State Networks \cite{maass2002real}. Furthermore, the present results demonstrate the potential of extending our RNN controller to other types of robots that use different sensory inputs.

\subsection{Capability of Evolving Complex Robotics Behavior}

One advantage of our method is the simple design of our fitness function. It only includes rewards for each visit of a novel reward and the completeness of each reward path plus a small penalty for repeated visits. We also tried with an additional reward term for the portion of time steps left before timeout, but finally excluded it from the fitness function. Instead, the time cost would be automatically influenced by rewarding non-repetitive reward path visits as demonstrated in Figure \ref{fig:evolved_trend}. It is not a fitness function only for a certain type of robot, because it is independent of robot properties (e.g., sensors, speed, motor structure, etc.). Therefore, it could be easily generalize to fit other task settings or robots. 

The key to performance in our experiments was the evolved weights into, within, and from the RNN. Generally a RNN has connections between internal neurons form a directed cycle. Arbitrary sequences of inputs could be processed by using the internal state of the RNN as the memory. RNNs have been applied to a broad range of domains such as terrain classification, motion prediction, and speech recognition \cite{zou2020terrain,kashyap2018recurrent,graves2013speech}. A reservoir-based approach, such as an LSM \cite{maass2002real}, can tractably harness such recurrence. 

Similar to an LSM, we also tried a reservoir-based approach to evolve only the output weights while keeping randomly initialized input and recurrent weights fixed throughout generations. In addition, we attempted to evolve both output weights along with input or recurrent weights (but not with both). However, their evolutionary performance could not generate fitness values as well as evolving all three types of weights. Evolving all three types of weights allow maximal utilization of neural activities to create the dynamics needed to solve a sequential memory task, such as the triple-T maze one. This process of evolving input, recurrent, and output weights could be readily transferred to other complex robotic settings. 

Because of the accurate representation of many popular robot designs in the Webots simulator, we could always run much faster than real-time (e.g., $\times30$ faster on average on our desktop with one GPU) to evolve for enough generations before applying a well-performed genotype to the RNN controller for real-world robotic navigation tasks. The power of using a detailed simulator, such as Webots, is that the evolved controller should transfer to the real e-puck with minimal adjustments.

\subsection{Comparison with Prior Work}

\subsubsection{Evolutionary robotics}
Evolutionary robotics is a method for building control system components or the morphology of a robot \cite{bongard2013evolutionary,nolfi2016evolutionary}. The biological inspiration behind this field is Darwin's theory of evolution, which was constructed with three principles. (1) \textit{Natural Selection}: genotypes that can be well adapted to their environments are more likely to survive and reproduce. (2) \textit{Heredity}: the fittest genotypes from the previous generation can be directly kept in the next generation; moreover, the new offspring in each generation is generated based on the selected genes from two parents. (3) \textit{Variation}: the new offspring goes through mutations and crossover with a certain probability and thus differs from both parents. 

Our method follows these three principles and falls into the common category of evolving the control system. However, what we evolved is novel compared to other work on evolutionary robotics. There has been work to evolve robots in cognitive tasks. For example, one group evolved virtual iCub humanoid robots to investigate the spontaneous emergence of emotions. Their populations were evolved to decide whether to ``keep'' or ``discard'' different visual stimuli \cite{pacella2017basic}. There has also been work on evolving robot controllers capable of navigating mazes. For example, Floreano's group evolved neural network controllers to navigate a maze without colliding into wall. Their neural networks were directly tested on a Khepera robot that had proximity sensors and two wheel motor system that was similar to the e-puck used in our present studies. Their evolved neural networks developed a direct mapping from the proximity sensors to the motors \cite{floreano2010evolution}. Our present work extends this prior work by evolving an RNN capable of navigating mazes, as well as demonstrating cognitive behavior.

Rather than evolving a direct mapping from sensors to motors, we instead evolved the weights from sensory inputs to the RNN, within the RNN, and from the RNN to the robot's motors as the genes for each genotype evolved in our network. As is discussed below, RNNs such as the ones we discuss here are neurobologically plausible and allow for comparisons with neuroscience and cognitive science data \cite{wang2018prefrontal, yang2019task}. Moreover, the RNN architecture is more generalizable to different types of robots working in complex scenarios (e.g., the triple T-maze with multiple rewards and landmarks), which results in optimal performance independent of any projection type. 

\subsubsection{Evolved RNNs}
Although there are only a few studies that have RNNs evolved directly in robotic experiments, evolving RNNs has been more frequently applied to virtual task settings. For example, \citet{akinci2019evolving} used either a steady-state genetic algorithm (SSGA) or an evolutionary strategy (ES) to evolve weights of the Long Short-Term-Memory (LSTM) network or RNN for the Lunar Lander game provided by the OpenAI gym \cite{brockman2016openai}. In their case, the ES developed more dynamic behavior than the SSGA, whereas the SSGA kept good genotypes and re-evaluated them with different configurations.

\citet{li2018opponent} evolved poker agents called ASHE with various types of evolutionary focus, such as learning diversified strategies from strong opponents, learning weakness from less competitive ones, learning opposite strategies at the same time, or a mix in-between. The genes in their GA covered all parameters in the estimators, including the LSTM weights, per-block initial states, and the estimator weights. 

A more biologically inspired example is related to the recent work by \citet{wieser2020eo}. Inspired by the neuroplasticity and functional hierarchies in the human neocortex, they proposed to use a network called EO-MTRNN to optimize neural timescales and restructure itself when training data underwent significant changes over time. 

All these related works have their unique perspectives that could inspire us to build a more robust and potentially faster evolutionary process for RNN systems in the future. For instance, we may consider to experiment with features in different evolutionary algorithms or co-evolve different neural regions which have different strategies or focus on a cognitive task.

\subsubsection{Comparison with NEAT/HyperNEAT robot controllers}
The evolutionary algorithms were utilized to evolve only weights in our RNN system and many other groups' work as mentioned earlier. Another popular evolutionary mechanism is NEAT, which has its unique feature of evolving the network topology together with the weights \cite{stanley2002evolving}. HyperNEAT extends NEAT by evolving connective CPPNs that generate patterns with regularities (e.g., symmetry, repetition, repetition with variation, etc.) \cite{stanley2009hypercube}. In the case of quadruped locomotion investigated by \citet{clune2009evolving}, HyperNEAT could evolve common gaits by exploiting the geometry to generate front-back, left-right, or diagonal symmetries. Our current model was tuned to have a fixed number of recurrent neurons since the first generation and have all-to-all connections between two layers or within the recurrent layer. It may be of interest to combine NEAT/HyperNEAT with topologies of RNNs to solve more complex problems and scenarios. For example, NEAT might be utilized in the beginning of the evolutionary process to efficiently derive a morphology for a more standard evolutionary algorithm to use in later generations. This hybrid approach has similarities to \citet{akinci2019evolving}. 

\subsubsection{Working memory}
Our evolved RNN encoded not only spatial information but also working memory to remember which paths had been traversed recently and which paths remained to be explored. Working memory helps to connect what happened earlier with what occurs later. It can be thought of as a general purpose memory system that can generalize, integrate and reason over information related to decision making or executive control \cite{diamond2013executive, vyas2020computation}. For example, \citet{yang2019task} trained single RNNs to perform 20 tasks simultaneously. Clustering of recurrent units emerged in their compositional task representation. Similar to biological neural circuits, their system could adapt to one task based on combined instructions for other tasks. Furthermore, individual units in their network exhibited different selectivity in various tasks. 

Working memory usually relies on the prefrontal cortex (PFC) for information maintenance and manipulation \cite{baddeley1994developments, eldreth2006evidence, smith1999storage}. \citet{wang2018prefrontal} investigated such brain functioning with a meta-reinforcement learning (meta-RL) system. Their model trained the weights of an RNN centered on PFC through a reward prediction error signal driven by dopamine (DA). This RNN ``learned to learn'', which means it had the ability to learn new tasks via its trained activation dynamics with no further tuning of its connection weights.

With further investigation and utilization of working memory, we also would like to have our evolved RNN generalize over multiple cognitive tasks and demonstrate cognitive functions observed in different brain regions.

\section{Conclusions}
In this paper, we introduced a recurrent neural network (RNN) model that linked the robot sensor values to its motor speed output. By evolving weights from sensory inputs to the RNN, within the RNN, and from the RNN to the robot's motors, the evolved network architecture achieved the goal of successfully performing a cognitive task that required spatial and working memory. The RNN population carried spatial information sufficient to localize robot in the triple T-maze. It also carried predictive information of which path robot intended on taking. Moreover, the robotic behavior was dependent on RNN dynamics rather than a sensor-to-motor mapping. Our method shows that complex robot behavior, similar to which being observed in animal models, can be evolved and realized in RNNs. 

\section*{Acknowledgment}
 This material is based upon work supported by the United States Air Force Award \#FA9550-19-1-0306. Any opinions, findings and conclusions or recommendations expressed in this material are those of the author(s) and do not necessarily reflect the views of the United States Air Force.

\bibliographystyle{plainnat}  
\bibliography{main.bib}

\end{document}